\documentclass[runningheads]{llncs}
\usepackage[T1]{fontenc}
\usepackage{graphicx}
\usepackage{amsmath}
\usepackage{amssymb}
\usepackage{multirow}
\begin{document}
\title{COLA-GEC: A Bidirectional Framework for \\Enhancing Grammatical Acceptability and \\Error Correction}
\author{Xiangyu Yang \and 
Xinying Qiu\textsuperscript{*} }

%\author{First Author\inst{1}\orcidID{0000-1111-2222-3333} \and
%Second Author\inst{2,3}\orcidID{1111-2222-3333-4444} \and
%Third Author\inst{3}\orcidID{2222--3333-4444-5555}}
%
\authorrunning{Yang et al.}

\institute{Department of Computer Science, School of Information Science and Technology\\
Guangdong University of Foreign Studies, Guangzhou, China\\ \email{xy.qiu@foxmail.com} 
}
%\institute{Princeton University, Princeton NJ 08544, USA \and
%Springer Heidelberg, Tiergartenstr. 17, 69121 Heidelberg, Germany
\maketitle              % typeset the header of the contribution
\begin{abstract}
Grammatical Error Correction (GEC) and grammatical acceptability judgment (COLA) are core tasks in natural language processing, sharing foundational grammatical knowledge yet typically evolving independently. This paper introduces COLA-GEC, a novel bidirectional framework that enhances both tasks through mutual knowledge transfer. First, we augment grammatical acceptability models using GEC datasets, significantly improving their performance across multiple languages. Second, we integrate grammatical acceptability signals into GEC model training via a dynamic loss function, effectively guiding corrections toward grammatically acceptable outputs. Our approach achieves state-of-the-art results on several multilingual benchmarks. Comprehensive error analysis highlights remaining challenges, particularly in punctuation error correction, providing insights for future improvements in grammatical modeling.

\keywords{GEC  \and COLA}
\end{abstract}
\footnotetext[1]{
	\textsuperscript{*}Corresponding author\\
}
\section{Introduction}

Grammatical Error Correction (GEC) and grammatical acceptability judgment (COLA) represent two fundamental tasks in natural language processing that directly address language proficiency needs of second-language learners \cite{lau2015}. GEC focuses on automatically identifying and correcting grammatical errors in text, while COLA aims to determine whether a given sentence adheres to the grammatical rules of a language. These capabilities have significant implications for educational applications and linguistic research.

In recent years, substantial progress has been made in both areas separately. For COLA, Warstadt et al. \cite{warstadt2019} established the Corpus of Linguistic Acceptability (CoLA) for English, inspiring similar resources for other languages including Italian \cite{trotta2021}, Russian  \cite{mikhailov2022}, Japanese \cite{someya2024}, and Chinese \cite{hu2023}. Meanwhile, GEC research has evolved from rule-based approaches to sophisticated neural models based primarily on Sequence-to-Sequence and Sequence-to-Edit architectures .

Despite their shared foundation in grammatical knowledge, these research areas have developed largely in isolation. The fundamental connection remains underexplored: a GEC system implicitly requires the ability to distinguish between grammatical and ungrammatical sentences, which is precisely the capability modeled by COLA systems. Similarly, COLA datasets typically contain minimal pairs of grammatical and ungrammatical sentences, while GEC datasets contain error-correction pairs that implicitly encode similar contrasts \cite{zhou2023}.

This paper addresses these gaps by establishing a bidirectional enhancement framework between GEC and COLA tasks. First, we enhance COLA models by incorporating GEC data, transforming error-correction pairs into acceptability judgment instances. Our GEC-enhanced COLA models achieve state-of-the-art performance across Chinese, English, German, and Arabic. Second, we integrate COLA into GEC models through an innovative architecture that evaluates candidate corrections and provides grammaticality signals to guide the GEC model toward more acceptable outputs. This approach achieves new state-of-the-art performance on multiple GEC datasets across languages. Finally, our comprehensive error analysis identifies specific grammatical phenomena—particularly punctuation errors—that present difficulties, providing valuable insights for future research in grammatical error correction systems. We provide our codes at: https://github.com/YXY2gdufs/colagec-project.

\section{Related Research}

\subsection{Grammatical Acceptability Judgment}

Research in grammatical acceptability judgment has developed along two main paths. First, large-scale annotated corpora have been established, beginning with Warstadt et al.'s \cite{warstadt2019} Corpus of Linguistic Acceptability (CoLA) for English, which collected examples from linguistics textbooks. This approach has been replicated for Italian \cite{trotta2021}, Russian  \cite{mikhailov2022}, Japanese \cite{someya2024}, and Chinese \cite{hu2023}, with most corpora containing binary judgments distinguishing grammatical from ungrammatical sentences. Modeling approaches initially utilized RNNs and LSTM swith various word embeddings, while recent work has leveraged pre-trained language models, with Mikhailov et al. investigating multilingual capabilities using XLM-R and RemBERT.

Second, researchers have developed methods to generate contrastive examples automatically. Warstadt et al. \cite{warstadt2020blimp} and Xiang et al. \cite{xiang2021climp} created BLiMP and CLiMP benchmarks respectively using grammatical templates with specific constraints. Nielsen \cite{nielsen2023scandeval} extended this approach by automatically generating ungrammatical sentences through word deletion or swapping for Scandinavian languages.

\subsection{Grammatical Error Correction}

Early GEC methods relied on rules and targeted classifiers, but struggled with complex errors involving multi-word dependencies. Current approaches primarily use Sequence-to-Sequence (Seq2Seq) or Sequence-to-Edit (Seq2Edit) architectures. The Seq2Seq approach treats GEC as a translation task, evolving from statistical machine translation to sophisticated neural architectures. Yuan and Felice\cite{yuan2013} pioneered the application of neural machine translation (NMT) to grammatical error correction (GEC). More recent innovations include the introduction of the Transformer architecture, dynamic masking to enhance generalization \cite{zhao2020}, the integration of pre-trained masked language models \cite{kaneko2020}, decoding interventions guided by external critics \cite{zhou2023}, and incorporating explanations into GEC systems to improve interpretability and performance \cite{fei2023}. Several studies have also investigated comparative performance of GEC techniques in learner writing, particularly in educational settings \cite{pham2025}.

Despite their shared foundation in grammatical knowledge, COLA and GEC research has developed separately. A GEC system implicitly requires the ability to distinguish between grammatical and ungrammatical sentences—precisely what COLA models do—while COLA datasets contain minimal pairs that parallel the error-correction pairs in GEC data. This suggests significant potential for bidirectional knowledge transfer, which we address in this paper.

\section{Methodology}
\subsection{Enhancing Grammatical Acceptability Models with GEC Data (G-Cola Model)}
Current grammatical acceptability judgment (COLA) models evaluate whether a sentence adheres to the grammatical rules of a language. We propose to enhance COLA corpora  by augmenting linguistic theory-based datasets with processed GEC datasets.

\subsubsection{Corpora Construction}
 For all languages, we use only the training set and the validation set of the GEC datasets to augment COLA corpora. The complete data statistics are provided in Table \ref{stats}. We discuss model training and implementation in Section 4.
 
\begin{table}[!htb]
        \centering
        \caption{Statistics of the Multilingual Corpus of Linguistic Acceptability judgments}
        \label{stats}
        \small 
        \begin{tabular}{@{}lccc@{}} 
            \hline
            Language      & Train   & Dev    & Test   \\ \hline
            English(en)   & 36,551  & 2,400  & 516    \\ 
            Chinese(zh)   & 16,772  & 2,928  & 931    \\ 
            Russian(ru)   & 7,869   & 1,483  & 2,341  \\ 
            Italian(it)   & 7,801   & 946    & 975    \\ 
            German(de)    & 19,621  & 2,803  & 402    \\ 
            French(fr)    & 500     & 521    & 521    \\ 
            Spanish(es)   & 500     & 321    & 322    \\ 
            Japanese(jp)  & 7,419   & 1,378  & 694    \\ 
            Arabic(ar)    & 14,330  & 2,413  & 313    \\ 
            Icelandic(is) & 500     & 1,194  & 1,194  \\ 
            Norwegian(no) & 8,030   & 1,000  & 1,000  \\ 
            Swedish(sw)   & 7,682   & 890    & 888    \\ \hline
        \end{tabular}
\end{table}

\paragraph{Chinese COLA Corpus}: We build upon CoLAC \cite{hu2023}, which contains 7,495 sentences from Chinese grammar textbooks, linguistics journals, and The Handbook of Chinese Linguistics. Following, we use crowd labels as ground truth (where "1" indicates grammatical acceptability and "0" indicates unacceptability). We incorporate sentence pairs from three Chinese GEC datasets: MuCGEC \cite{zhang2022-muc}, FCGEC \cite{xu2022}, and NaSGEC \cite{zhang2023nasgec}, yielding 1,059, 7,862, and 4,115 high-quality sentence pairs, respectively. For each pair, corrected sentences are labeled as grammatically acceptable (1) and error-containing sentences as unacceptable (0). The resulting corpus contains 6,072 sentences from CoLAC and 10,700 sentences from GEC datasets.

\paragraph{English COLA Corpus}: We apply a similar approach to the original CoLA corpus \cite{warstadt2019}, which contains 9,594 sentences from syntax textbooks, verb classification resources \cite{levin1993}, and papers on syntactic movement. We augment this with filtered sentence pairs from W\&I+LOCNESS \cite{bryant2019}, FCE \cite{yannakoudakis2011}, and NUCLE \cite{dahlmeier2013}, selecting 10,312, 9,706, and 9,855 sentence pairs, respectively. The enhanced corpus contains 8,551 sentences from CoLA and 28,000 sentences from GEC datasets.

\paragraph{German and Arabic COLA Corpora}: For German and Arabic, we started with theoretical linguistics textbook examples collected by Zhang et al. \cite{zhang2024}. We enhanced the German corpus with sentence pairs from the Falko and MERLIN learner corpora, and the Arabic corpus with sentence pairs from QALB-2014 and QALB-2015 datasets.

\paragraph{Multilingual COLA Corpus}: Building on the MELA corpus, which covers 10 languages, we incorporate additional languages and GEC data. We enhance the Japanese portion with JCoLA  \cite{someya2024}, add Norwegian \cite{jentoft2023} and Swedish \cite{volodina2021} acceptability corpora, and augment multiple languages with their respective GEC datasets. The final corpus covers 12 languages with standardized binary labels.

\subsection{COLA-Guided Training for Grammatical Error Correction (G-Cola GEC Model)}
We now present our approach for integrating grammatical acceptability signals into GEC training, addressing limitations in current GEC models.

\subsubsection{Framework Overview:}
As illustrated in Figure 1, our framework integrates a COLA model to evaluate and guide the GEC training process through the following components:

\begin{itemize}
\item We train a BART-based Seq2Seq GEC model that transforms ungrammatical input sentences into corrected versions
\item We add GEC data to enhance pretrained COLA model using corpora discussed in Section 3.1. 
\item We implement a Dynamic loss based on Label Confidence Weighted Learning (LCWL) mechanism \cite{qiu2024} that incorporates COLA model judgments into the GEC model's loss function. We name our GEC model G-Cola GEC.
\item We plug in our G-Cola GEC model to the GEC-di model (which is a dual-critc model achieving SOTA performances on English and Chinese GEC tasks). The G-Cola GEC-di model performs inference on GEC test sets.
\end{itemize}

\begin{figure*}
\centering
\includegraphics[scale=0.05]{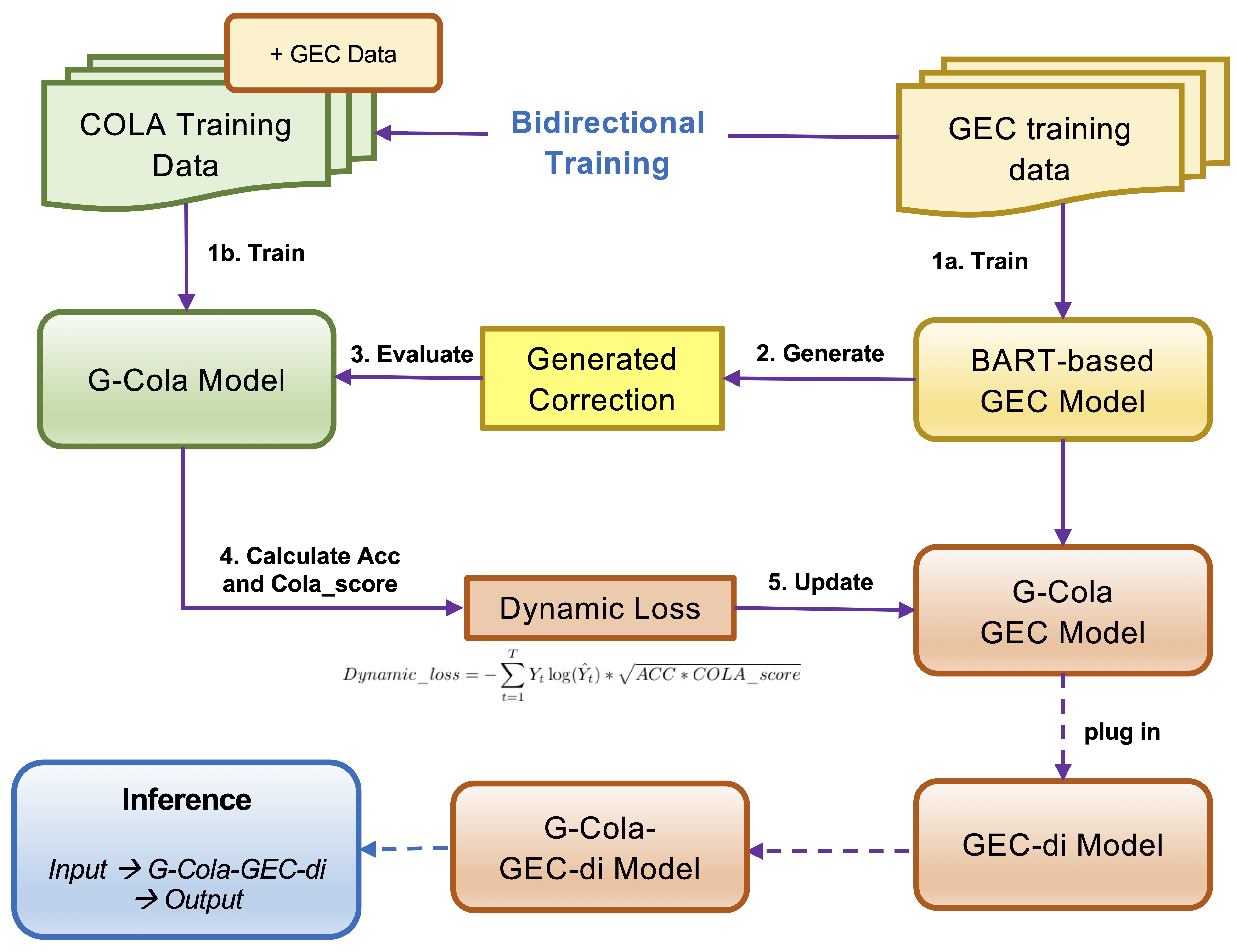}
\caption{Bidirectional Framework of COLA-GEC} 
\label{fig1}
\end{figure*}

\subsubsection{Incorporating COLA judgements into GEC training:}
The COLA model evaluates corrected sentences from the GEC model by producing scores for two categories: "0" (grammatically unacceptable) and "1" (grammatically acceptable). We transform these using a sigmoid function:

\begin{equation}
COLA\_score = \frac{1}{1 + e^{-(Logits\_0-Logits\_1)}}
\end{equation}
where $\text{Logits}_0$ and $\text{Logits}_1$ represent scores for categories 0 and 1 respectively. When $\text{Logits}_0 > \text{Logits}_1$ (indicating grammatical unacceptability), the score increases, signaling grammatical errors in the correction.

To integrate COLA judgments into GEC training, we adapt the Label Confidence Weighted Learning (LCWL) approach \cite{qiu2024}. We modify the  GEC cross-entropy loss to incorporate COLA scores:
\begin{equation}
Dynamic\_loss = -\sum_{t=1}^{T} Y_t \log(\hat{Y}_t) * \sqrt{Acc * COLA\_score}
\end{equation}
where  $T$ is sequence length, $Y_t$ is the ground truth at time step t, $\hat{Y}_t$ is the model's prediction, and $Acc$ is the COLA model's accuracy. This formulation weighs the loss according to grammatical acceptability, with higher penalties for grammatically problematic corrections.

Through this integrated approach, we develop G-Cola GEC, a grammatical confidence-guided error correction model that leverages acceptability judgments  to improve correction quality across multiple languages.

\section{Training and Implementation}

\subsubsection{G-Cola Training and Evaluation}

We implement two training strategies for grammatical acceptability judgment:

\paragraph{Monolingual Models}: We fine-tune RoBERTa-based models for English, Chinese, and German using their respective GEC-enhanced datasets. Each model consists of a pre-trained language model encoder followed by a classification layer predicting binary grammatical acceptability. We name our monolingual COLA model G-Cola-EN, G-Cola-CN, and G-Cola-DE.
\paragraph{Multilingual Model}: We fine-tune XLM-R using our combined corpus of 12 languages, with Chinese, English, German and Arabic enhanced by GEC data. This leverages cross-lingual transfer learning, enabling knowledge sharing across languages. We name this model G-Cola-Multi.

We compare our two strategies with MELA  \cite{zhang2024} , which is the current SOTA COLA model, and vanilla RoBERTa or XLM-R based COLA model without GEC data augmentation. All models are evaluated using accuracy (ACC) and Matthews correlation coefficient (MCC).

\subsubsection{G-Cola GEC Training and Evaluation}
Our GEC models were trained on substantial corpora with high error rates. The Chinese model used Lang8 (590K sentences, 89.5\% error rate) and HSK (157K sentences, 60.8\% error rate) for pretraining, followed by fine-tuning on FCGEC, MuCGEC, and NaSGEC datasets. The English model pretrained on Clang8 (2.37M sentences, 57.8\% error rate) and fine-tuned on W\&I+LOCNESS, FCE, and NUCLE datasets. The German model used Falko (11K sentences, 65.5\% error rate) for pretraining and MERLIN (10K sentences, 78.7\% error rate) for fine-tuning.
\paragraph{Chinese Implementation}: Our Chinese GEC model used BART-large-Chinese as the base architecture. Following Zhou et al. \cite{zhou2023}, we employed a two-stage training strategy:
\begin{enumerate}
\item Pre-training on Lang8 corpus (split 8:2 for training/validation)
\item Fine-tuning on HSK, MuCGEC-dev, FCGEC train and dev, and NaSGEC-Exam train and dev. We name this model G-CoLA-GEC-CN.
\end{enumerate}

The resulting G-CoLA-GEC-CN model was further combined with GECdi-CN (a dual-critic framework \cite{zhou2023} ) to create G-CoLA GECdi-CN, which was evaluated on MuCGEC and FCGEC test sets. We compare our performances with GEC di \cite{zhou2023} , MrGEC \cite{liu2024}, and EPOGEC \cite{liang2025}.

\paragraph{English Implementation}: We used BART-large for English GEC and implemented a three-stage training approach:
\begin{enumerate}
\item Pre-training on Clang8 corpus
\item First fine-tuning on W\&I+LOCNESS
\item Second fine-tuning on FCE, NUCLE, and BEA-19-dev to adapt to diverse error types. We name this model G-CoLA-GEC-EN.
\end{enumerate}
The resulting G-CoLA GEC-EN was also combined with Zhou's dual-critic framework \cite{zhou2023} to create G-CoLA GECdi-EN, evaluated on CoNLL-14 and BEA-19 test sets. We compare our model's performance with GEC di \cite{zhou2023} , MultimodelGEC\cite{fang2023}, EPOGEC \cite{liang2025}, Unsupervised GEC \cite{cao2023unsupervised}.

\paragraph{German Implementation}: For low-resource German, we based our model on Hugging Face's Bart-German We pretrained on Falko and finetuned on MERLIN. This model was enhanced with our G-Cola-DE model because it performs better than G-Cola-Multi on German dataset. Furthermore, since there is no dual-critic GEC model\cite{zhou2023} for German, we trained GEC-Di-DE as follows:
\begin{enumerate}
\item We train a language model critic (LM-Critic) based on German-GPT2, fine-tuned on 9,000 sentences from German Wikipedia articles on social and cultural topics
\item We train a target-side grammatical error detection critic (Target GED Critic) based on Bart-German, to identify specific error types using ERRANT-computed error annotations
\item We combined the above two step to implement Gecdi-DE (i.e. adapting the dual-critic mocel \cite{zhou2023} for German).
\end{enumerate}
The resulting Gecdi-DE model was combined with the G-Cola-DE and evaluated on the Falko-MERLIN test set. We name this model G-CoLA GECdi-DE and compare with MultimodelGEC \cite{fang2023}, and WikiGerman GEC \cite{boyd2018}.

All models were evaluated using language-specific metrics: ChERRANT for Chinese and ERRANT for English and German, with performance measured through Precision (P), Recall (R), and F0.5 score. 

\section{Results and Analysis}
\subsection{G-Cola Model Results}

We distinguish between baseline models (CoLA) trained without GEC corpus data and our enhanced models (G-CoLA) incorporating GEC corpus data. As shown in Table \ref{tab:cola_performance}, G-CoLA models consistently outperform their CoLA counterparts across all languages, with substantial improvements in both accuracy and MCC metrics. The Chinese G-CoLA model improved by 4.13\% in accuracy and 0.59 in MCC, while the English model showed improvements of 2.59\% and 2.11 respectively. For lower-resource languages, German and Arabic G-CoLA models demonstrated similar gains, confirming the effectiveness of our approach across diverse languages. Table \ref{tab:cola_performance} indicates that our multilingual G-CoLA model surpasses MELA (current state-of-the-art) across all languages, with an average MCC improvement of 2.22. Notably, performance improved not only for languages with directly added GEC data, but also for typologically related languages, suggesting beneficial cross-linguistic transfer effects, particularly within language families.

% Combine Tables 1, 2, 3 into one comprehensive table
\begin{table}[h]
\centering
\footnotesize
\caption{G-Cola Model Performance Comparison}
\label{tab:cola_performance}
\begin{tabular}{lcccccccc}
\hline
\multirow{2}{*}{\textbf{Model}} & \multicolumn{2}{c}{\textbf{Chinese}} & \multicolumn{2}{c}{\textbf{English}} & \multicolumn{2}{c}{\textbf{German}} & \multicolumn{2}{c}{\textbf{Arabic}} \\
\cline{2-9}
& ACC & MCC & ACC & MCC & ACC & MCC & ACC & MCC \\
\hline
MELA(2024) & - & 54.94 & - & 60.64 & - & 26.72 & - & 14.12 \\
CoLA & 81.30 & 56.61 & 85.51 & 63.49 & 62.90 & 37.74 & 60.67 & 36.30 \\
G-CoLA & \textbf{85.43} & \textbf{57.20} & \textbf{88.10} & \textbf{65.60} & \textbf{66.42} & \textbf{41.81} & \textbf{64.90} & \textbf{42.24} \\
G-CoLA-Multi & 77.95 & 55.37 & 83.2 & 61.93 & 42.92 & 29.55 & 40.1 & 18.33 \\
\hline
\end{tabular}
\end{table}

% Combine Tables 4, 5, 7 into one GEC results table
\begin{table}[h]
\centering
\scriptsize
\caption{G-Cola GECdi Performance Across Languages}
\label{tab:gec_performance}
\begin{tabular}{l ccccccccccccccc}
\hline
\multirow{3}{*}{\textbf{Model}} & \multicolumn{3}{c}{\textbf{Chinese }}& \multicolumn{3}{c}{\textbf{Chinese }}& \multicolumn{3}{c}{\textbf{English }}& \multicolumn{3}{c}{\textbf{English }} & \multicolumn{3}{c}{\textbf{German }} \\
&  \multicolumn{3}{c}{(MuCGEC)} &\multicolumn{3}{c}{(FCGEC)} &  \multicolumn{3}{c}{(CoNLL-14)} &  \multicolumn{3}{c}{(BEA-19)} &  \multicolumn{3}{c}{(Falko-MERLIN)} \\
\cline{2-16}
& Pre & Rec & F0.5 &Pre & Rec & F0.5 & Pre & Rec & F0.5 & Pre & Rec & F0.5 & Pre & Rec & F0.5 \\
\hline
GECdi (2023) & 56.74 & 31.0 & 48.61 &  - & - & - & 79.2 & 46.8 & 69.6  & 77.4 & 59.9 & 73.1   & - & - & - \\
MultiModal (2023) &  - & - & - &  - & - & - &  75 & 53.2 & 69.3 &  77.1 & 66.7 & 74.8 &  78.50 & 68.40 & 76.30\\
MrGEC (2024) & 56.37 &  29.12 & 47.47 & 65.71 & 37.78 & 57.22 &  - & - & -& - & - & - & - & - & -\\
EPOGEC (2025) & - & - & -  & 66.67 & \textbf{41.93} & \textbf{59.63} & 76.71 & 52.56 & 70.26 & 78.16 & 68.07 & 75.91 & - & - & - \\
Unsupervised (2023)  & - & - & - & - & - & - & 75 & \textbf{53.8} & 69.6 & \textbf{78.8} & \textbf{68.5} & \textbf{76.5} & - & - & - \\
WikiGerman(2018) & - & - & - & - & - & - & - & - & - & - & - & - & 45.22 & 51.99 & 29.73 \\
\textbf{G-CoLA GECdi} & \textbf{57.12} & \textbf{33.9} & \textbf{49.91} & \textbf{67.1} & 39.05 & 58.6 & \textbf{79.55} & 49.04 & \textbf{70.84} & 77.33 & 61.67 & 74.03 & 71.93 & 62.47 & 70.80\\
\hline
\end{tabular}
\end{table}

\subsection{G-Cola GECdi Results}

We evaluated our G-CoLA GECdi models across multiple languages as shown in Table \ref{tab:gec_performance}. For Chinese, G-CoLA GECdi-CN achieved state-of-the-art results on MuCGEC, with the highest scores across all metrics. On FCGEC, our model achieved the highest precision (67.11\%) but slightly lower F0.5 than EPOGEC. For English, G-CoLA GECdi-EN achieved the highest precision (79.55\%) and F0.5 (70.84\%) on CoNLL-14, exceeding GEC di by 0.35 and 1.24 points respectively, though its recall was lower than some competitors. On BEA-19, our model performed competitively but below Unsupervised GEC. For German, G-CoLA GECdi-DE substantially outperformed WikiGerman GEC (70.80 vs 29.73 F0.5) but remained below MultimodelGEC (76.30 F0.5), which leverages additional visual information our text-only approach doesn't use. These results demonstrate that incorporating grammatical acceptability signals consistently enhances GEC performance across languages, validating our bidirectional framework's effectiveness.

\subsection{Error Analysis}

%修改
Our model's lower performance on BEA-19 versus CoNLL-14 stems from different error compositions, as shown in Table \ref{tab:punc}. BEA-19 contains significantly more challenging error types: punctuation errors (16.73\% vs 7.60\%) and uncommon errors (15.69\% vs 1.50\%), totaling 32.4\% versus 9.1\% in CoNLL-14. Additional differences include higher orthographic errors (8.03\% vs 3.81\%) in BEA-19 and higher word choice errors (14.4\%) in CoNLL-14.

\begin{table}[]
\begin{center}
\label{tab:punc}
\caption{English GEC Test Dataset Percentage of Error Types}
\begin{tabular}{ccc}
\hline
Error types & CONLL-14 test & BEA-19 test \\
\hline
Wci & 14.4\% & - \\
Rloc- & 6.00\% & - \\
Nn & 6.80\% & 4.07\% \\
SVA & 4.60\% & 2.28\% \\
PUNCT & \textbf{7.60\%} & \textbf{16.73\%} \\
OTHER & \textbf{1.50\%} & \textbf{15.69\%} \\
PREP & 11.70\% & 8.33\% \\
DET & 11.40\% & 10.41\% \\
ORTH & 3.81\% & 8.03\% \\
VERB: TENSE & 4.95\% & 5.43\% \\
VERB & 6.43\% & 5.09\% \\
SPELL & 6.09\% & 4.63\% \\
VERB: FORM & 3.10\% & - \\
NOUN: INFL & - & 2.89\% \\
\hline
\end{tabular}
\end{center}
\end{table}

We tested our hypothesis that COLA models struggle with punctuation errors using the BEA-19 validation set (913 punctuation errors, 672 other uncommon errors). Three experiments were conducted: (1) evaluating G-CoLA GECdi-EN on the complete set; (2) removing punctuation/other errors individually or combined; and (3) targeted fine-tuning.

As Table \ref{tab:experiment_results} shows, removing punctuation errors improved F0.5 from 72.25 to 74.40, while removing both error types yielded 75.92 F0.5. Targeted fine-tuning showed minimal improvement (73.11 F0.5), indicating that specialized approaches beyond simple data augmentation are needed for these challenging error categories.

\begin{table}[h]\textbf{}
    \begin{center}
\caption{Result of error analysis experiment}
\label{tab:experiment_results}
\begin{tabular}{ccccc}
\hline
Model                                  & Dev                                                                      & Pre                    & Rec                    & F0.5                   \\ \hline
G-CoLA GECdi-EN                     & BEA-19 Dev                                                               & 76.19                  & 60.01                  & 72.25                  \\ \hline
G-CoLA GECdi-EN                     & -PUNCT                                                                   & 77.93                  & 63.94                  & 74.40                   \\
G-CoLA GECdi-EN                     & -OTHER                                                                   & 77.03                  & 62.53                  & 73.96                  \\
\multirow{2}{*}{G-CoLA GECdi-EN}    & \multirow{2}{*}{\begin{tabular}[c]{@{}c@{}}-PUNCT\\ -OTHER\end{tabular}} & \multirow{2}{*}{\textbf{79.88}} & \multirow{2}{*}{\textbf{64.37}} & \multirow{2}{*}{\textbf{75.92}} \\
                                    &                                                                          &                        &                        &                        \\ \hline
G-CoLA GECdi-EN 
\\for PUNCT and OTHER & BEA-19 Dev                                                               & 76.55                  & 61.37                  & 73.11                  \\ \hline
\end{tabular}
\label{tab:error_analysis}
\end{center}
\end{table}

\subsection{Ablation Study}

%表4-14

To further validate the impact of grammatical acceptability models enhanced with GEC data (G-Cola) on GEC performance, we conducted comprehensive ablation experiments comparing five model configurations: BART-based GEC, BART-based GEC + G-Cola, GECdi, GECdi+Cola, and GECdi+G-Cola. As shown in Table 8, incorporating grammatical acceptability models consistently improved performance across all standard GEC test sets in Chinese and English. Adding the baseline CoLA model to GEC systems produced notable gains, while using our G-CoLA model (enhanced with GEC data) yielded even greater improvements. For instance, on the CoNLL-14 test set, adding G-CoLA to the base GEC model improved F0.5 from 54.08 to 59.17, while integrating G-CoLA with GECdi further increased F0.5 to 70.84. Similar patterns were observed across all datasets and languages, with particularly substantial gains on the Chinese test sets (MuCGEC, FCGEC, and NaSGEC-Exam). These results confirm that grammatical acceptability models enhanced with error-correction data provide significant and consistent benefits to GEC systems, validating our bidirectional enhancement framework.

\begin{table}[h]\textbf{}
    \begin{center}
\caption{Result of ablation study experiment}
\label{tab:experiment_results}
\begin{tabular}{ccccccc}
\hline
Test                                                                        & Metric                   & GEC       & \multicolumn{1}{c|}{+G-CoLA} & GECdi                & +CoLA                & +G-CoLA              \\ \hline
\multirow{3}{*}{CONLL14 test}                                               & Pre                  & 63.02                & \multicolumn{1}{c|}{\textbf{67.33}}   & 79.32                & 79.13                & \textbf{79.55}                \\
& Rec       & 41.26                & \multicolumn{1}{c|}{\textbf{48.33}}   & 48.55                & 46.82                & \textbf{49.04 }               \\
    & F0.5                 & 54.08                & \multicolumn{1}{c|}{\textbf{59.17}}   & 70.11                & 69.57                & \textbf{70.84}                \\ \hline
\multirow{3}{*}{BEA-19 test}                                                & Pre                  & 64.41                & \multicolumn{1}{c|}{\textbf{64.53}}   & 77.26                & 77.23                & \textbf{77.33}                \\
&Rec & 51.11                & \multicolumn{1}{c|}{\textbf{53.47}}   & 60.52                & 59.5                 & \textbf{61.67}                \\
& F0.5                 & 61.12                & \multicolumn{1}{c|}{\textbf{62.97}}   & 73.55                & 73.1                 & \textbf{74.03}                \\ \hline
\multirow{3}{*}{MuCGEC test}                                                & Pre                  & 43.73                & \multicolumn{1}{c|}{\textbf{45.85}}   & 56.83                & 56.76                & \textbf{57.12}                \\
    & Rec                  & 28.34                & \multicolumn{1}{c|}{\textbf{31.77}}   & 33.71                & 31.08                & \textbf{33.9}                 \\
     & F0.5                 & 39.75                & \multicolumn{1}{c|}{\textbf{42.93}}   & 49.77                & 48.7                 & \textbf{49.91}                \\ \hline
\multirow{3}{*}{FCGEC test}                                                 & Pre                  & 48.04                & \multicolumn{1}{c|}{\textbf{50.95}}   & 66.7                 & 66.19 \textbf{}               & \textbf{67.11}                \\
     & Rec                  & 30.05                & \multicolumn{1}{c|}{\textbf{34.69}}   & 38.54                & 36.91                & \textbf{39.05}                \\
     & F0.5                 & 42.75                & \multicolumn{1}{c|}{\textbf{46.93}}   & 57.77     & 57.08                & \textbf{58.6}                 \\ \hline
\multirow{3}{*}{\begin{tabular}[c]{@{}c@{}}NaSGEC-Exam\\ test\end{tabular}} & Pre                  & 45.62                & \multicolumn{1}{c|}{\textbf{53.67}}   & 62.7                 & 62.37                & \textbf{63.92}                \\
& Rec                  & 21.94                & \multicolumn{1}{c|}{\textbf{25.13}}   & 31.54                & 30.4                 & \textbf{32.1}                 \\
    & F0.5                 & 40.25                & \multicolumn{1}{c|}{\textbf{67.33}}   & 79.32                & 79.13                & \textbf{79.55}                \\ \hline
\multicolumn{1}{l}{}                                                        & \multicolumn{1}{l}{} & \multicolumn{1}{l}{} & \multicolumn{1}{l}{}         & \multicolumn{1}{l}{} & \multicolumn{1}{l}{} & \multicolumn{1}{l}{}
\end{tabular}
\end{center}
\end{table}
\textbf{}

\newpage
\section{Conclusions}

This paper presented COLA-GEC, a novel bidirectional framework that enhances both grammatical acceptability judgment (COLA) and grammatical error correction (GEC) through cross-task knowledge transfer. We demonstrated that augmenting COLA models with GEC data significantly improves grammatical acceptability judgment performance across multiple languages, with consistent gains in both monolingual and multilingual settings. Conversely, incorporating COLA signals into GEC training through dynamic loss mechanism substantially improved GEC performance across Chinese, English, and German benchmarks. Our error analysis revealed specific challenges with punctuation errors and certain grammatical phenomena, providing valuable insights for future research. The ablation studies further confirmed the effectiveness of our bidirectional approach, showing that each component contributes meaningfully to overall performance gains. Future work will focus on addressing punctuation error challenges and developing specialized models for uncommon grammatical phenomena.
%\section{Appendix}


\section{Appendix}

% \usepackage{multirow}
\begin{table}[]

    \begin{center}
\caption{G-COLA Monolingual Dataset Statistics }
\label{tab1}
\begin{tabular}{ccccc}
\hline
Language                     & Source       & Train & Dev  & Test \\ \hline
\multirow{4}{*}{Chinese(Zh)} & CoLAC        & 6072  & 492  & 931  \\
                             & MuCGEC       & 700   & 359  & 0    \\
                             & FCGEC        & 6100  & 1862 & 0    \\
                             & NaSGEC       & 3900  & 215  & 0    \\ \hline
\multirow{4}{*}{English(En)} & CoLA         & 8551  & 527  & 516  \\
                             & W\&I+LOCNESS & 9700  & 612  & 0    \\
                             & FCE          & 9100  & 606  & 0    \\
                             & NUCLE        & 9200  & 655  & 0    \\ \hline
\multirow{3}{*}{German(De)}  & CoLA-De      & 500   & 402  & 402  \\
                             & Falko        & 10021 & 1205 & 0    \\
                             & MERLIN       & 9100  & 1196 & 0    \\ \hline
\multirow{3}{*}{Arabic(Ar)}  & CoLA-Ar      & 500   & 313  & 313  \\
                             & QALB-2014    & 13500 & 1960 & 0    \\
                             & QALB-2015    & 330   & 140  & 0    \\ \hline
\end{tabular}
\end{center}
\end{table}
%\newpage
\begin{table}[h]
\centering
\caption{Chinese G-Cola GEC Corpus Statistics}
\begin{tabular}{cccc}
\hline
Stage & Corpus & Number of Sentences & Error Percentage \\
\hline
\multirow{3}{*}{Pretrain} & Lang8-all & 590059 & 89.5\% \\
& HSK-all & 156870 & 60.8\% \\
& MuCGEC-dev & 1125 & 95.1\% \\
\hline
\multirow{5}{*}{Finetune} & FCGEC-train & 36341 & 55.3\% \\

& FCGEC-dev & 2000 & 55.1\% \\

& NaSGEC-Exam-train & 4000 & 67.9\% \\

& NaSGEC-Exam-dev & 1000 & 72.3\% \\
\hline
\multirow{3}{*}{Test} & MuCGEC-test & 5938 & 92.2\% \\

& FCGEC-test & 3000 & 54.5\% \\

& NaSGEC-Exam-test & 2000 & 68.6\% \\
\hline
\end{tabular}
\end{table}

\begin{table}[h]
\centering
\caption{English G-Cola GEC Corpus Statistics}
\begin{tabular}{cccc}
\hline
Stage & Corpus & Number of Sentences & Error Percentage \\
\hline
\multirow{1}{*}{Pretrain} & Clang8-all & 2372119 & 57.8\% \\
\hline
\multirow{4}{*}{Finetune} & FCE-all & 34490 & 62.6\% \\
& NUCLE-all & 57151 & 38.2\% \\
& W\&I+LOCNESS-all & 34308 & 66.3\% \\
& BEA-19-dev & 4384 & 65.2\% \\
\hline
\multirow{2}{*}{Test}& BEA-19-test & 4477 & - \\
& CoNLL-14-test & 1312 & 72.3\% \\
\hline
\end{tabular}
\end{table}

\begin{table}[h]
\centering
\caption{German G-Cola GEC Corpus Statistics}
\begin{tabular}{cccc}
\hline
Stage & Corpus & Number of Sentences & Error Percentage \\
\hline
Pretrain & Falko-all & 11335 & 65.5\% \\
\hline
Finetune & MERLIN-all & 10395 & 78.7\% \\
\hline
Test & Falko-MERLIN & 2337 & 76.6\% \\
\hline

\end{tabular}
\end{table}

\end{document}